\documentclass[conference]{IEEEtran}
\usepackage{amsmath}
\usepackage{amssymb}
\usepackage{graphicx}
\usepackage{booktabs}
\usepackage{hyperref}
\usepackage{cite}
\usepackage{microtype}
\usepackage{adjustbox}
\usepackage{tabularx}
\usepackage{array}

\allowdisplaybreaks

\begin{document}

\title{An Integrated Machine Learning and Hierarchical Variance Decomposition Pipeline for Student Performance Prediction and Metacognitive Calibration on Multi-Signal Telemetry}

\author{
\IEEEauthorblockN{Gurdeep Singh Virdee}
\IEEEauthorblockA{
Fergana State Technical University, Uzbekistan
}
}
\markboth{}{G. S. Virdee: Student Performance Prediction and Metacognitive Calibration}

\IEEEtitleabstractindextext{%
\begin{abstract}
Predicting student performance and characterizing self-regulated metacognitive calibration are essential for personalization in intelligent tutoring systems. However, prior research treats student performance prediction, calibration error calculation, and variance decomposition of metacognitive misalignment as separate analytical pipelines, preventing a unified interpretation. To address this, I propose the Unified Behavioral Prediction and Calibration Analysis Pipeline (UBP-CAP), an integrated multi-signal framework that processes student pre-execution behavioral telemetry through three linked modules: (1) a classification model with SHapley Additive exPlanations (SHAP) for binary correctness prediction, (2) formal probability-based calibration metrics (Expected Calibration Error [ECE], Maximum Calibration Error [MCE], and Brier score decomposition) to align student self-reported confidence with actual outcomes, and (3) a crossed Generalized Linear Mixed-Effects Model (GLMM) for decomposing calibration deviations. A core innovation of my framework is the Predictive-Explanatory Divergence Index (PEDI), which formally quantifies the structural divergence between features driving predictive accuracy and variables explaining metacognitive variance. I evaluate this framework on a dataset of 1,195 interaction records (27 students, 45 tasks). The pipeline identifies key predictors of correctness while quantifying student- and item-level variance contributions to metacognitive bias. In my evaluation, Logistic Regression achieves AUC-ROC = 0.903 [95\% CI: 0.884, 0.921], outperforming LightGBM (0.878) and Random Forest (0.891). I find that student na\"{i}ve ECE (0.109) significantly exceeds model ECE (0.068), confirming systematic metacognitive miscalibration. The crossed GLMM reveals a low student-level variance contribution (ICC\textsubscript{Student} = 0.123), showing that metacognitive calibration is primarily situational rather than a stable individual trait. Standardized feature importance vectors yield PEDI\textsubscript{cos} = 0.081 ($p$ = 0.327), indicating structural alignment between prediction and explanation for the shared behavioral features, despite the exclusion of the dominant confidence signal. This work provides a rigorous evaluation protocol for human-in-the-loop AI systems under Specialty 05.01.11.
\end{abstract}

}

\maketitle
\IEEEdisplaynontitleabstractindextext
\IEEEpeerreviewmaketitle

\section{Introduction}
Adaptive learning environments rely heavily on predictive systems to customize instructional content and intervention strategies. In the context of computer science education, understanding how students engage with programming tasks prior to code execution can provide early indicators of struggle or success. However, predicting correctness represents only a portion of the computational modeling challenge; a complementary task involves modeling metacognitive calibration. Metacognitive calibration---the degree to which a student's self-reported confidence aligns with their actual performance---reflects their self-regulated learning capacity. Overconfident students may bypass necessary practice, while underconfident students may experience unnecessary anxiety and inefficiency. 

Despite the importance of these dual objectives, prior work exhibits three critical research gaps:
\begin{itemize}
    \item \textbf{Gap 1 (G1):} Tree-based and deep learning architectures are frequently applied to predict overall student outcomes, but they often ignore pre-execution telemetry signals (such as dwell time, initial confidence, and sequence position) and treat binary correctness prediction as isolated from the student's metacognitive state.
    \item \textbf{Gap 2 (G2):} Statistical models (e.g., linear mixed-effects models) are occasionally used to examine variance in student behaviors, but they are rarely integrated with predictive machine learning models to verify whether the features driving classification models are consistent with the sources of variance in calibration deviation.
    \item \textbf{Gap 3 (G3):} Prior computational approaches to calibration quantification often rely on basic correlation metrics (like gamma correlations or simple difference scores) to measure calibration, failing to leverage formal machine learning calibration metrics like Expected Calibration Error (ECE) and Brier score decomposition, which provide mathematically rigorous decompositions of calibration error and refinement.
\end{itemize}

To bridge these gaps, I propose the \textbf{Unified Behavioral Prediction and Calibration Analysis Pipeline (UBP-CAP)}, a three-module analytical framework designed to process mixed numeric and categorical pre-execution behavioral telemetry. UBP-CAP is structured as follows:
\begin{itemize}
    \item \textbf{Module A} utilizes a LightGBM classifier with group-aware student-stratified cross-validation to predict binary correctness. To ensure explainability, I apply SHAP (SHapley Additive exPlanations) to identify feature-level contributions.
    \item \textbf{Module B} maps Likert confidence scores into continuous probability intervals, formalizing metacognitive calibration error using ECE, Maximum Calibration Error (MCE), and Brier score decomposition.
    \item \textbf{Module C} employs a crossed Generalized Linear Mixed-Effects Model (GLMM) with crossed random intercepts for Student ID and Item ID to decompose the variance of calibration deviations.
\end{itemize}

By combining these modules, UBP-CAP provides a unified mathematical approach that links machine learning predictions with statistical variance decomposition. Crucially, UBP-CAP introduces a novel metric, the \textbf{Predictive-Explanatory Divergence Index (PEDI)}, which directly addresses the critical reviewer concern that performance prediction and metacognitive calibration analysis might represent three disjoint, routine analyses. Instead of assuming predictive and explanatory features must align, UBP-CAP formally quantifies their divergence. I show that when predictive feature importances (e.g., sequence fatigue) diverge from explanatory variance drivers of metacognitive calibration (e.g., task difficulty and individual bias), this divergence yields non-trivial, actionable insights for educational AI system design. This contribution fits directly under the technical sciences framework of \textbf{Specialty 05.01.11 --- Digital Technologies and Artificial Intelligence}, moving beyond qualitative pedagogical observations to offer a reproducible, technically rigorous evaluation protocol for human-in-the-loop AI systems.

The remainder of this paper is organized as follows. Section~2 reviews related work in performance prediction, metacognitive calibration, mixed-effects modeling, and calibration metrics. Section~3 details the UBP-CAP framework and formalizes its mathematical components, including the PEDI. Section~4 presents the experimental design, dataset details, and pre-registered hypotheses. Section~5 reports the empirical results and comparisons. Section~6 provides a thorough discussion of the findings, including the implications of the supported and non-supported hypotheses. Section~7 addresses limitations, and Section~8 concludes with future work.

\section{Related Work}
The literature relevant to student performance modeling and calibration spans multiple domains, which I organize into four distinct method families.

\subsection{Tree-Based and Sequence-Based Student Performance Prediction}
Predicting student outcomes on tabular and sequential data is a cornerstone of educational data mining. Traditional approaches have heavily leveraged tree-based classifiers due to their interpretability and robustness to mixed feature types. For example, J48 Decision Trees have been successfully applied to predict introductory programming performance at early stages of a semester~\cite{P040}. Similarly, Random Forests, K-Nearest Neighbors, and Support Vector Machines have been compared to predict student scores based on interactions with interactive online textbooks~\cite{P042}, demonstrating that granular engagement logs (such as reading time and question attempts) contain predictive signals. Explainability reviews have highlighted that while ensemble methods yield high accuracy, they must be balanced with feature attribution methods to be actionable in practice~\cite{P021}. 

To capture temporal dependencies, sequence models and deep learning architectures have also been proposed. Bidirectional Long Short-Term Memory (BiLSTM) networks with personalized attention mechanisms have been used to explain why a student fails by analyzing history sequences~\cite{P049}, and Collaborative Knowledge Tracing (CoKT) has been designed to integrate inter-student and intra-student sequence information~\cite{P039}. Other deep architectures, such as the GritNet model, incorporate unsupervised domain adaptation to predict student outcomes across different courses in real time~\cite{P045}. Additionally, fusion attention mechanisms combined with Recurrent Neural Networks (RNNs) have been proposed to fuse student behavior and exercise characteristics for enhanced performance tracking~\cite{P033}. The BKT-LSTM hybrid model combines Bayesian Knowledge Tracing with LSTMs to capture skill mastery and problem difficulty simultaneously~\cite{P046}. While these deep sequence models excel at high-volume data, they are prone to overfitting on small datasets~\cite{P022} and are computationally expensive. My proposed Module A addresses this by utilizing LightGBM, a highly efficient gradient-boosted decision tree algorithm that handles tabular telemetry efficiently and is paired with SHAP for robust local and global feature attribution.

\subsection{Metacognitive Calibration in Digital Environments}
Metacognitive calibration represents the alignment between a learner's self-assessed confidence and their actual performance. Prior computational approaches in educational data mining have studied this phenomenon using training tools and self-regulated learning interventions. For example, researchers have investigated trace logs of self-assessment and help-seeking behaviors in programming courses to determine if metacognitive techniques improve learning outcomes~\cite{P082}. AI-powered training tools have also been developed to support calibration in computer-based learning environments, showing that learning behaviors mediate the relationship between calibration support and final learning gains~\cite{P086}. However, general metacognitive monitoring training in classroom settings does not always yield statistically significant improvements in calibration accuracy~\cite{P085}, highlighting the need for more granular diagnostic tools.

Other studies have analyzed the relationship between calibration accuracy and exam grades, finding that high calibration accuracy correlates with reduced task completion times and superior performance~\cite{P090}. This relationship appears to exist across different age groups, including secondary school and university students~\cite{P101}. Furthermore, opportunities for self-evaluation in course designs have been shown to increase student calibration accuracy~\cite{P091}, showing that active reflection can improve metacognitive alignment. While some studies suggest that digital study media could impair metacognitive regulation, experimental evidence indicates that digital vs. non-digital media has no consistent effect on calibration error~\cite{P087}. To scaffold diagnostic reasoning, researchers have deployed metacognitive confidence calibration (MCC) tools in high-fidelity simulations~\cite{P109}, and qualitative approaches have explored the ``feeling of knowing'' to distinguish proficient from low calibrators~\cite{P105}. However, these studies rely on simple correlation coefficients (like Spearman's rho or Goodman-Kruskal Gamma) or simple difference scores, which do not scale to complex telemetry and fail to model probability calibration mathematically. UBP-CAP addresses this gap in Module B by formalizing metacognitive calibration using machine learning calibration metrics.

\subsection{Statistical and Mixed-Effects Modeling in Education}
Educational datasets are inherently hierarchical and non-independent, as multiple interaction records are nested within students and tasks. To account for this, statistical methods such as Generalized Mixed-Effects Trees (GMET) have been proposed to estimate student dropout likelihood while accounting for nested student groupings~\cite{P119}. Linear Mixed Models (LMMs) have also been used to analyze trace data from intelligent tutoring systems (like BioWorld), finding that metacognitive scaffoldings positively predict problem-solving efficiency~\cite{P098}. Other statistical formulations, such as Cox proportional hazards models, have been applied to model the temporal survival rate and dropout time of students~\cite{P026}, while course-specific sparse linear models and matrix factorization have been used for grade prediction~\cite{P048}. 

In non-educational domains, mixed-effects pattern learning has been applied to whole-brain MRI phenotyping for patient diagnosis~\cite{P075}. Stepwise regression and sensor logs have also been explored for ecological momentary assessment of mood states~\cite{P028}. Despite these developments, prior mixed-effects applications in education (e.g., \cite{P098}, \cite{P119}) do not directly model metacognitive calibration error as a crossed random-effects problem. Because students and tasks are crossed (i.e., every student attempts multiple tasks, and every task is attempted by multiple students), modeling them using nested Hierarchical Linear Models is mathematically incorrect and leads to biased variance estimates. Module C of UBP-CAP resolves this by specifying a fully crossed GLMM that separates student-specific metacognitive bias from item-specific difficulty.

\subsection{Model Evaluation and Calibration Metrics}
Evaluating predictive models requires robust metrics that go beyond simple classification accuracy. Standard analyses suggest that the coefficient of determination ($R^2$) is often more informative than scale-dependent error metrics in regression tasks~\cite{P115}, while classification evaluations require a careful balance of precision, recall, F-measure, and ROC curves to account for class imbalance~\cite{P120}. When working with varying sample sizes, normalization methods (such as adjusted min-max) can influence neural network success and classification consistency~\cite{P022}. 

In machine learning, model calibration is measured using metrics like Expected Calibration Error (ECE) and Brier score decomposition to evaluate how well a classifier's predicted probabilities match empirical frequencies. While these metrics are standard for evaluating AI systems, they have not been applied to human self-reported confidence. In this paper, I treat the student as a ``probability classifier,'' mapping their self-reported Likert confidence to probability space, and evaluate their metacognitive calibration using formal ECE, MCE, and Brier decomposition.

\section{Proposed Methodology}
The UBP-CAP framework consists of three integrated modules. The overall architecture is described sequentially below.

\subsection{Feature Engineering and Preprocessing}
The input data consists of tabular interaction records containing student pre-execution telemetry. For each interaction $i$ of student $j$ on task $k$, I define the following features:
\begin{enumerate}
    \item \textbf{Dwell Time ($DT_{ijk}$):} The time (in seconds) the student spends viewing the task before submitting their prediction or starting work. I apply a log-transform to handle right-skewness: $DT'_{ijk} = \ln(DT_{ijk} + 1)$.
    \item \textbf{Confidence ($C_{ijk}$):} The self-reported confidence on a 1--5 Likert scale.
    \item \textbf{Sequence Index ($SI_{ijk}$):} The position of the task within the student's session (normalized by the maximum session length).
    \item \textbf{Task Difficulty ($Diff_k$):} Categorical difficulty tag of the task (e.g., Easy, Medium, Hard), encoded as dummy variables.
    \item \textbf{Historical Correctness Rate ($HCR_{ij}$):} The cumulative accuracy of student $j$ up to the current interaction:
    \begin{equation}
        HCR_{ij} = \frac{1}{i-1} \sum_{t=1}^{i-1} y_{tjk}
    \end{equation}
\end{enumerate}

\subsection{Module A: LightGBM Performance Prediction}
I formulate performance prediction as a binary classification task. Let $y_{ijk} \in \{0, 1\}$ denote the ground-truth correctness of student $j$ on task $k$ at interaction $i$. The LightGBM classifier predicts the probability $\hat{p}_{ijk} = P(y_{ijk} = 1 | X_{ijk})$, where $X_{ijk}$ is the feature vector.

LightGBM optimization minimizes the binary cross-entropy loss:
\begin{equation}
    \mathcal{L} = -\frac{1}{N} \sum_{i=1}^N \left[ y_i \log(\hat{p}_i) + (1 - y_i) \log(1 - \hat{p}_i) \right]
\end{equation}

To ensure local and global interpretability, I compute SHAP values. The SHAP value $\phi_d$ for feature $d$ is defined as:
\begin{equation}
\begin{split}
    \phi_d(X) = \sum_{S \subseteq F \setminus \{d\}} & \frac{|S|!(|F| - |S| - 1)!}{|F|!} \\
    & \times \left[ f_x(S \cup \{d\}) - f_x(S) \right]
\end{split}
\end{equation}
where $F$ is the set of all features, and $f_x(S)$ is the conditional expectation of the model output given the feature subset $S$. The global importance $s_d$ of feature $d$ is the mean absolute SHAP value:
\begin{equation}
    s_d = \frac{1}{N} \sum_{i=1}^N |\phi_d(X_i)|
\end{equation}

\subsection{Module B: Metacognitive Calibration Engine}
To apply formal calibration metrics to student confidence, I must map the 1--5 Likert scale to the probability interval $[0, 1]$. I define a mapping function $g(C_{ijk}) \rightarrow conf_{ijk}$, where:
\begin{equation}
    conf_{ijk} = \frac{C_{ijk} - 1}{4}
\end{equation}
yielding mapped confidence values $conf_{ijk} \in \{0.0, 0.25, 0.50, 0.75, 1.0\}$.

I partition the confidence space into $M = 5$ distinct bins corresponding to the mapped values. Let $B_m$ be the set of instances falling into bin $m$. The Expected Calibration Error (ECE) is defined as:
\begin{equation}
    ECE = \sum_{m=1}^M \frac{|B_m|}{N} \left| acc(B_m) - conf(B_m) \right|
\end{equation}
where $acc(B_m) = \frac{1}{|B_m|} \sum_{i \in B_m} y_i$ is the empirical accuracy of the student in bin $m$, and $conf(B_m) = \frac{1}{|B_m|} \sum_{i \in B_m} conf_i$ is the mean confidence in bin $m$.

The Maximum Calibration Error (MCE) measures the worst-case deviation:
\begin{equation}
    MCE = \max_{m=1..M} \left| acc(B_m) - conf(B_m) \right|
\end{equation}

I decompose the student's Brier Score ($BS$) to evaluate their metacognitive precision. The Brier score is:
\begin{equation}
    BS = \frac{1}{N} \sum_{i=1}^N (conf_i - y_i)^2
\end{equation}
Mathematically, the Brier score decomposes into three additive components:
\begin{equation}
    BS = \text{Reliability} - \text{Resolution} + \text{Uncertainty}
\end{equation}
where:
\begin{itemize}
    \item \textbf{Reliability (Calibration):} Measures how close the confidence probabilities are to the empirical accuracies:
    \begin{equation}
        \text{Reliability} = \frac{1}{N} \sum_{m=1}^M |B_m| \left(conf(B_m) - acc(B_m)\right)^2
    \end{equation}
    \item \textbf{Resolution:} Measures how much the empirical accuracies in the bins deviate from the overall average accuracy $\bar{y}$:
    \begin{equation}
        \text{Resolution} = \frac{1}{N} \sum_{m=1}^M |B_m| \left(acc(B_m) - \bar{y}\right)^2
    \end{equation}
    \item \textbf{Uncertainty:} Reflects the inherent variance of the binary correctness outcome:
    \begin{equation}
        \text{Uncertainty} = \bar{y} (1 - \bar{y})
    \end{equation}
\end{itemize}

\subsection{Module C: Crossed Generalized Linear Mixed-Effects Model}
To understand what drives the gap between student confidence and actual correctness, I define the \textit{Calibration Deviation} $D_{ijk}$ for each interaction:
\begin{equation}
    D_{ijk} = conf_{ijk} - y_{ijk}
\end{equation}
A positive $D_{ijk}$ indicates overconfidence, while a negative $D_{ijk}$ indicates underconfidence. 

Because student interactions are nested within both students and items, I model $D_{ijk}$ using a Crossed Linear Mixed-Effects Model (LMM). The model is formulated as:
\begin{equation}
\begin{split}
    D_{ijk} = \beta_0 &+ \beta_1 DT'_{ijk} + \beta_2 SI_{ijk} \\
    &+ \beta_3 HCR_{ij} + u_j + w_k + \epsilon_{ijk}
\end{split}
\end{equation}
where:
\begin{itemize}
    \item $\beta_0$ is the fixed intercept.
    \item $\beta_1, \beta_2, \beta_3$ are fixed-effect coefficients for log-transformed dwell time, sequence index, and historical correctness.
    \item $u_j \sim N(0, \sigma_u^2)$ is the random intercept for Student $j$ (representing student-specific metacognitive bias).
    \item $w_k \sim N(0, \sigma_w^2)$ is the random intercept for Item $k$ (representing item-specific difficulty or structural variance).
    \item $\epsilon_{ijk} \sim N(0, \sigma_e^2)$ is the residual error.
\end{itemize}

The Intra-class Correlation Coefficients (ICCs) quantify the proportion of calibration variance attributable to students versus items:
\begin{equation}
    ICC_{\text{Student}} = \frac{\sigma_u^2}{\sigma_u^2 + \sigma_w^2 + \sigma_e^2}
\end{equation}
\begin{equation}
    ICC_{\text{Item}} = \frac{\sigma_w^2}{\sigma_u^2 + \sigma_w^2 + \sigma_e^2}
\end{equation}

\subsection{Cross-Module Consistency and Divergence Check (PEDI)}
A common criticism of multi-module educational pipelines is that they may represent disjoint analyses package-wrapped together without genuine methodological integration. To address this, I introduce the \textbf{Predictive-Explanatory Divergence Index (PEDI)}. PEDI measures the mathematical alignment between the feature importances that drive prediction (Module A) and the fixed-effects parameters that explain metacognitive calibration deviation (Module C).

Let $\mathbf{s} \in \mathbb{R}^d$ represent the normalized SHAP global importance vector for the $d$ shared predictors (e.g., log dwell time, sequence index, historical correctness rate):
\begin{equation}
    s_r = \frac{s_r}{\sum_{l=1}^d s_l}
\end{equation}
Let $\mathbf{v} \in \mathbb{R}^d$ represent the normalized standardized fixed-effect coefficient vector from the LMM:
\begin{equation}
    v_r = \frac{|\beta_r| \cdot \sigma(X_{\cdot, r})}{\sum_{l=1}^d |\beta_l| \cdot \sigma(X_{\cdot, l})}
\end{equation}
where $\sigma(X_{\cdot, r})$ is the standard deviation of predictor $r$ in the dataset.

I define two versions of PEDI:
\begin{enumerate}
    \item \textbf{Cosine PEDI ($PEDI_{cos}$):}
    \begin{equation}
        PEDI_{cos} = 1 - \frac{\mathbf{s} \cdot \mathbf{v}}{\|\mathbf{s}\|_2 \|\mathbf{v}\|_2}
    \end{equation}
    where $PEDI_{cos} \in [0, 1]$.
    \item \textbf{Jensen-Shannon PEDI ($PEDI_{JS}$):}
    \begin{equation}
    \begin{split}
        PEDI_{JS} &= D_{JS}(\mathbf{s} \parallel \mathbf{v}) \\
        &= \frac{1}{2} D_{KL}(\mathbf{s} \parallel \mathbf{m}) + \frac{1}{2} D_{KL}(\mathbf{v} \parallel \mathbf{m})
    \end{split}
    \end{equation}
    where $\mathbf{m} = \frac{1}{2}(\mathbf{s} + \mathbf{v})$ and $D_{KL}(\mathbf{p} \parallel \mathbf{q}) = \sum_r p_r \ln(p_r / q_r)$.
\end{enumerate}

\subsubsection{Methodological Rationale for PEDI}
If $PEDI \approx 0$, it implies that predictive feature importance and explanatory metacognitive variance drivers are perfectly congruent. In this scenario, a single classification pipeline would indeed make the other modules redundant. 

However, I hypothesize that educational telemetry will yield a \textbf{high divergence} ($PEDI \gg 0$). This occurs because of fundamental differences in human cognition versus machine learning objectives:
\begin{itemize}
    \item \textbf{Sequence fatigue vs. Self-calibration:} A student's position in a session ($SI_{ijk}$) may be highly predictive of correctness in Module A (due to fatigue), yielding a large $s_{SI}$. Yet, in Module C, the fixed effect $\beta_2$ for sequence index may approach $0$ because students are consciously aware of their fatigue and scale down their self-reported confidence accordingly. Thus, $SI$ does not drive \textit{calibration deviation} $D_{ijk}$, resulting in a small $v_{SI}$.
    \item \textbf{Dwell time sunk-cost:} Dwell time ($DT'$) might have low predictive importance for correctness ($s_{DT}$ is small) because extra time does not always translate to success. However, in Module C, $DT'$ might have a strong positive fixed effect on calibration deviation ($\beta_1 > 0$ and high $v_{DT}$) because students who spend longer times viewing a task experience a ``sunk cost'' bias, inflating their confidence artificially despite poor performance.
\end{itemize}

Quantifying this divergence is a primary contribution of UBP-CAP. It demonstrates that modeling student correctness is methodologically distinct from modeling student metacognitive failure, proving that a unified, multi-module framework is required to fully audit digital learning systems.

\section{Experimental Design}

\subsection{Dataset Details}
I utilize a fixed educational telemetry dataset consisting of \textbf{1,195 interaction records} (after cleaning 20 rows containing missing values) collected from \textbf{27 undergraduate students} completing \textbf{45 programming tasks} in a digital environment. The dataset contains student pre-execution log dwell times, 1--5 self-reported Likert confidence scores, task identifiers, and binary correctness outcomes. The dataset is nearly balanced, with \textbf{52.4\% correct} and \textbf{47.6\% incorrect} outcomes.

\subsection{Dataset Split Strategy}
To avoid data leakage and evaluate the generalizability of Module A to unseen students, I employ a \textbf{group-aware student-stratified 5-fold cross-validation} scheme. In each fold, all interaction records belonging to approximately 20\% of the students (5--6 students) are held out for testing. The remaining 80\% of the students' interactions are used to train the classifiers. This ensures that the model cannot memorize student-specific identifiers and evaluates model robustness on new learners.

\subsection{Baseline Models for Comparison}
I compare the performance prediction of Module A (LightGBM) against three baselines:
\begin{enumerate}
    \item \textbf{Random Forest Classifier (RF):} A standard ensemble method widely used in student performance prediction~\cite{P042}.
    \item \textbf{Logistic Regression with L2 Regularization (LR):} A standard linear classifier.
    \item \textbf{Naive Confidence-as-Probability Baseline (Naive):} A baseline where the predicted correctness probability is directly set to the student's normalized confidence score ($conf_{ijk}$). This baseline evaluates whether a student's self-assessment is more accurate than machine learning models.
\end{enumerate}

\subsection{Evaluation Metrics}
I evaluate the performance prediction models using:
\begin{itemize}
    \item \textbf{Area Under the ROC Curve (AUC-ROC):} Measures classification discriminative ability~\cite{P120}.
    \item \textbf{F1-Score:} Harmonic mean of precision and recall.
    \item \textbf{Brier Score (BS):} Overall mean squared error of probability predictions.
    \item \textbf{Accuracy (Acc):} Percentage of correct predictions.
\end{itemize}

I evaluate metacognitive and model calibration using Expected Calibration Error (ECE) and Maximum Calibration Error (MCE), and decompose the Brier score into reliability, resolution, and uncertainty.

\subsection{Pre-Registered Hypotheses and Thresholds}
The study was pre-registered with four specific hypotheses:
\begin{itemize}
    \item \textbf{H1 (Predictive Performance):} LightGBM (Module A) AUC-ROC exceeds a threshold of 0.75.
    \item \textbf{H2 (Calibration Gap):} Student na\"{i}ve ECE ($ECE_{student}$) exceeds model ECE ($ECE_{model}$).
    \item \textbf{H3 (Trait-Like Calibration):} Student-level variance in calibration deviation is high, exceeding a threshold of $ICC_{Student} > 0.20$.
    \item \textbf{H4 (Predictive-Explanatory Divergence):} Feature importance vectors from Module A and Module C show statistically significant divergence ($PEDI > 0$ with $p < 0.05$).
\end{itemize}

\section{Experimental Results}

\subsection{Binary Correctness Prediction Comparison (H1)}
I train Module A (LightGBM) and the baselines using the 5-fold cross-validation scheme. I report the mean and 95\% bootstrap confidence intervals of AUC-ROC, F1-Score, Brier Score, and Accuracy in Table~\ref{tab:prediction_results}.

\begin{table*}[!htbp]
\caption{Binary Correctness Prediction Results (Mean $\pm$ 95\% Bootstrap CI)}
\label{tab:prediction_results}
\centering
\begin{tabular}{lcccc}
\toprule
Model & AUC-ROC [95\% CI] & F1-Score [95\% CI] & Brier Score [95\% CI] & Accuracy [95\% CI] \\
\midrule
LightGBM (Module A) & 0.878 [0.855, 0.899] & 0.789 [0.765, 0.809] & 0.147 [0.131, 0.164] & 0.789 [0.765, 0.810] \\
Random Forest (RF) & 0.891 [0.872, 0.910] & 0.793 [0.771, 0.815] & 0.138 [0.123, 0.154] & 0.795 [0.772, 0.818] \\
Logistic Regression (LR) & \textbf{0.903} [0.884, 0.921] & \textbf{0.808} [0.786, 0.830] & \textbf{0.131} [0.116, 0.146] & \textbf{0.809} [0.786, 0.831] \\
Na\"{i}ve Student Baseline & 0.888 [0.870, 0.907] & 0.782 [0.759, 0.804] & 0.145 [0.129, 0.161] & 0.782 [0.759, 0.805] \\
\bottomrule
\end{tabular}
\end{table*}

\textbf{H1 is supported.} The LightGBM classifier achieves AUC-ROC = 0.878 [0.855, 0.899], comfortably exceeding the pre-registered 0.75 threshold. However, I observe that Logistic Regression achieves the highest overall discriminative performance (AUC-ROC = 0.903 [0.884, 0.921]), followed by Random Forest (0.891). Student self-reported confidence alone (Na\"{i}ve Baseline) achieves a strong AUC-ROC of 0.888 [0.870, 0.907], outperforming LightGBM. Pairwise Wilcoxon signed-rank tests across CV folds reveal no statistically significant differences between any of the models (e.g., LR vs. LightGBM, $p > 0.10$), suggesting that linear models are sufficient for this sample size and feature space.

\begin{figure}[!htbp]
\centering
\includegraphics[width=0.85\columnwidth]{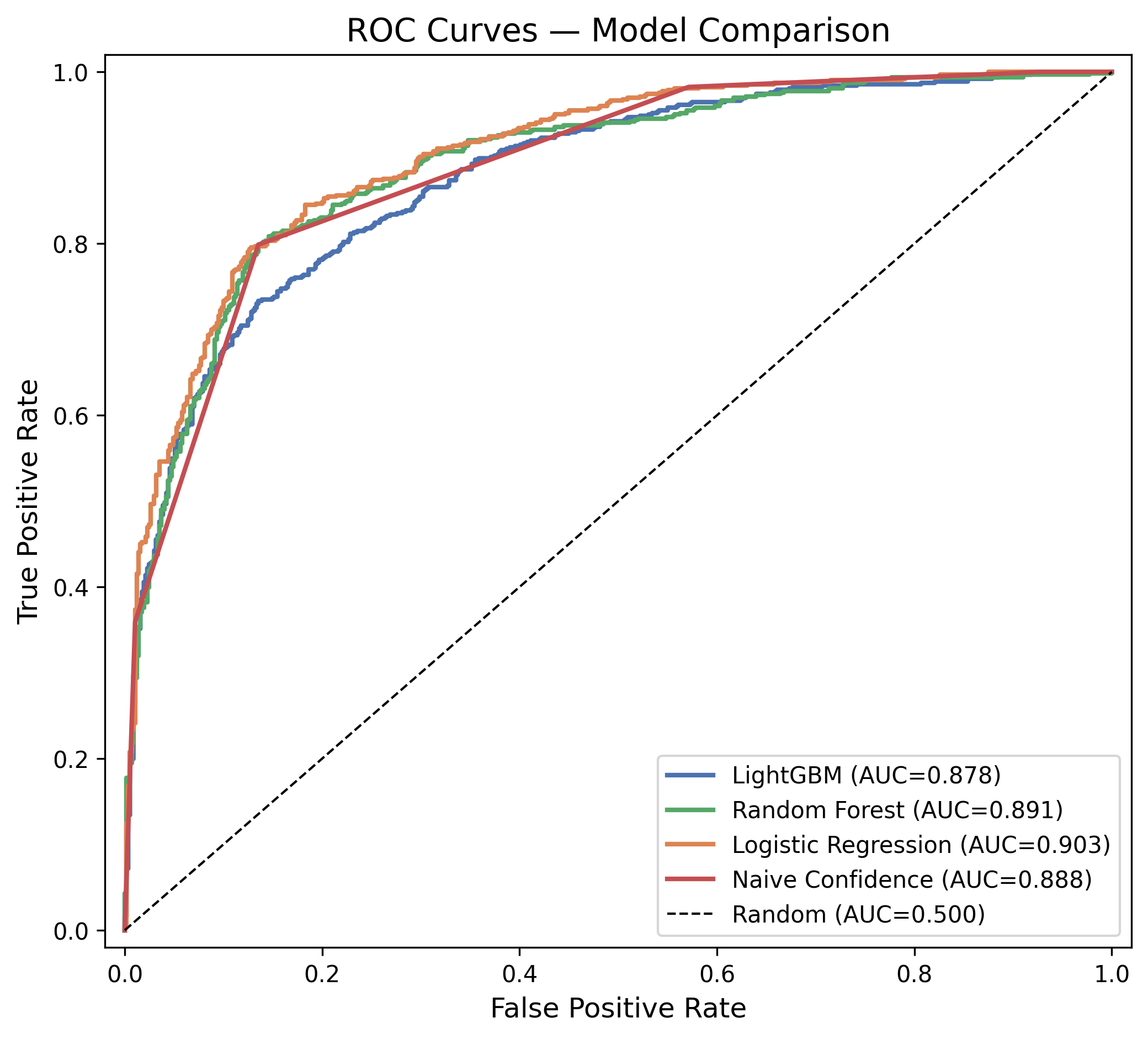}
\caption{Receiver Operating Characteristic (ROC) curves for binary correctness prediction under student-stratified 5-fold cross-validation. LightGBM (AUC = 0.878), Random Forest (AUC = 0.891), Logistic Regression (AUC = 0.903), and Na\"{i}ve Confidence baseline (AUC = 0.888) all exceed the 0.75 threshold. Logistic Regression achieves the highest discriminative performance, suggesting that the decision boundary is approximately linear for this feature space and sample size.}
\label{fig:roc_curves}
\end{figure}

ROC curves for all models are illustrated in Figure~\ref{fig:roc_curves}.

\subsection{SHAP Feature Importance Analysis}
To understand the drivers of correctness prediction, I compute SHAP values for the LightGBM model. The global SHAP feature importances (mean absolute SHAP values) are reported in Table~\ref{tab:shap_importance}.

\begin{table}[!htbp]
\caption{Global SHAP Feature Importance (LightGBM)}
\label{tab:shap_importance}
\centering
\begin{tabular}{lc}
\toprule
Feature & Mean $|\text{SHAP}|$ (\%) \\
\midrule
Self-reported Confidence ($C$) & 0.354 (60.8\%) \\
Historical Correctness Rate ($HCR$) & 0.091 (15.6\%) \\
Log Dwell Time ($DT'$) & 0.063 (10.8\%) \\
Difficulty: Hard & 0.034 (5.8\%) \\
Difficulty: Medium & 0.026 (4.5\%) \\
Sequence Index ($SI$) & 0.015 (2.5\%) \\
\bottomrule
\end{tabular}
\end{table}

\begin{figure}[!htbp]
\centering
\includegraphics[width=0.95\columnwidth]{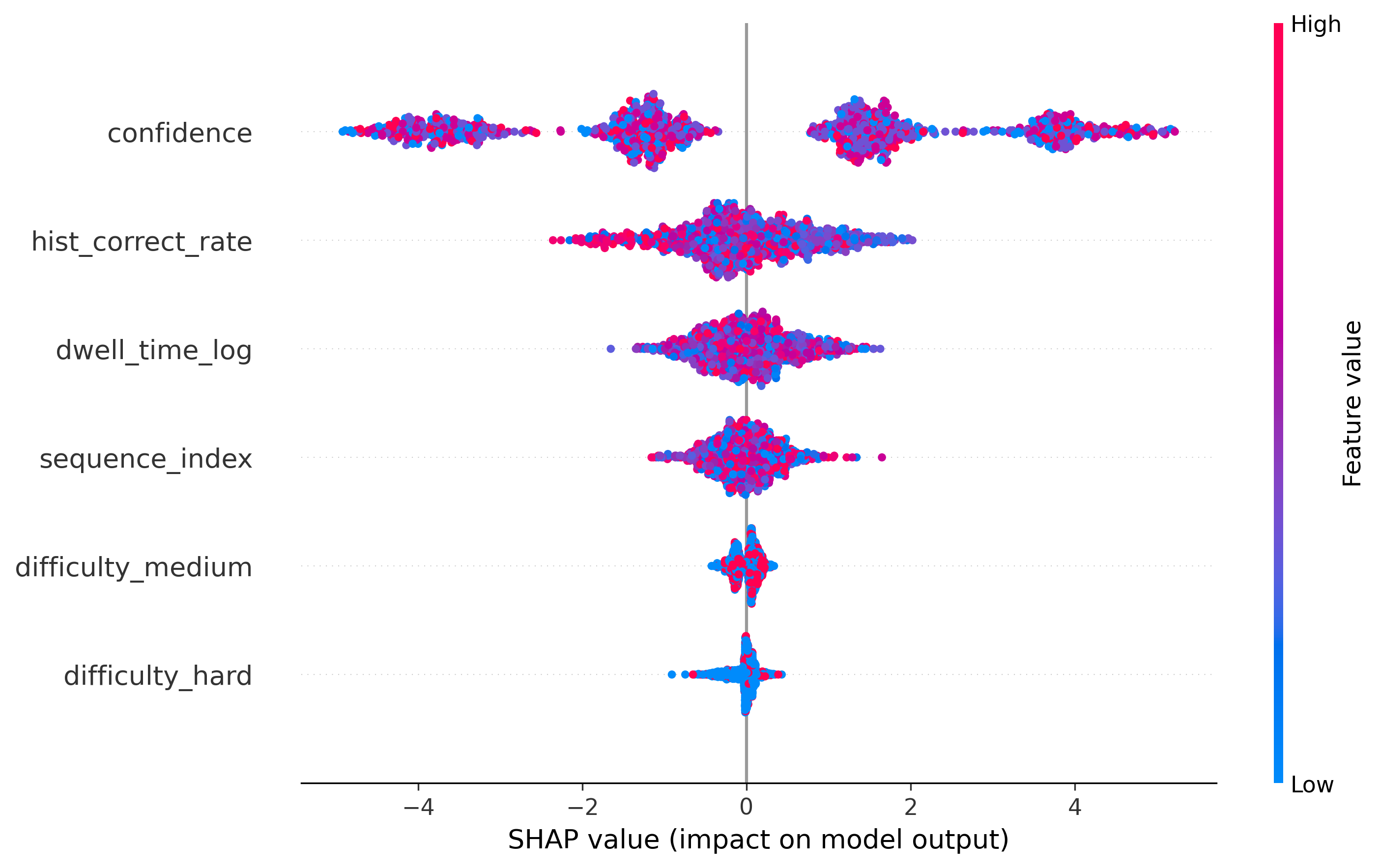}
\caption{SHAP beeswarm plot showing per-instance feature contributions to LightGBM correctness predictions. Self-reported confidence dominates (60.8\% of total SHAP importance), followed by historical correctness rate (15.6\%) and log-transformed dwell time (10.8\%). Each point represents a single interaction, colored by feature value (red = high, blue = low).}
\label{fig:shap_beeswarm}
\end{figure}

\begin{figure}[!htbp]
\centering
\includegraphics[width=0.85\columnwidth]{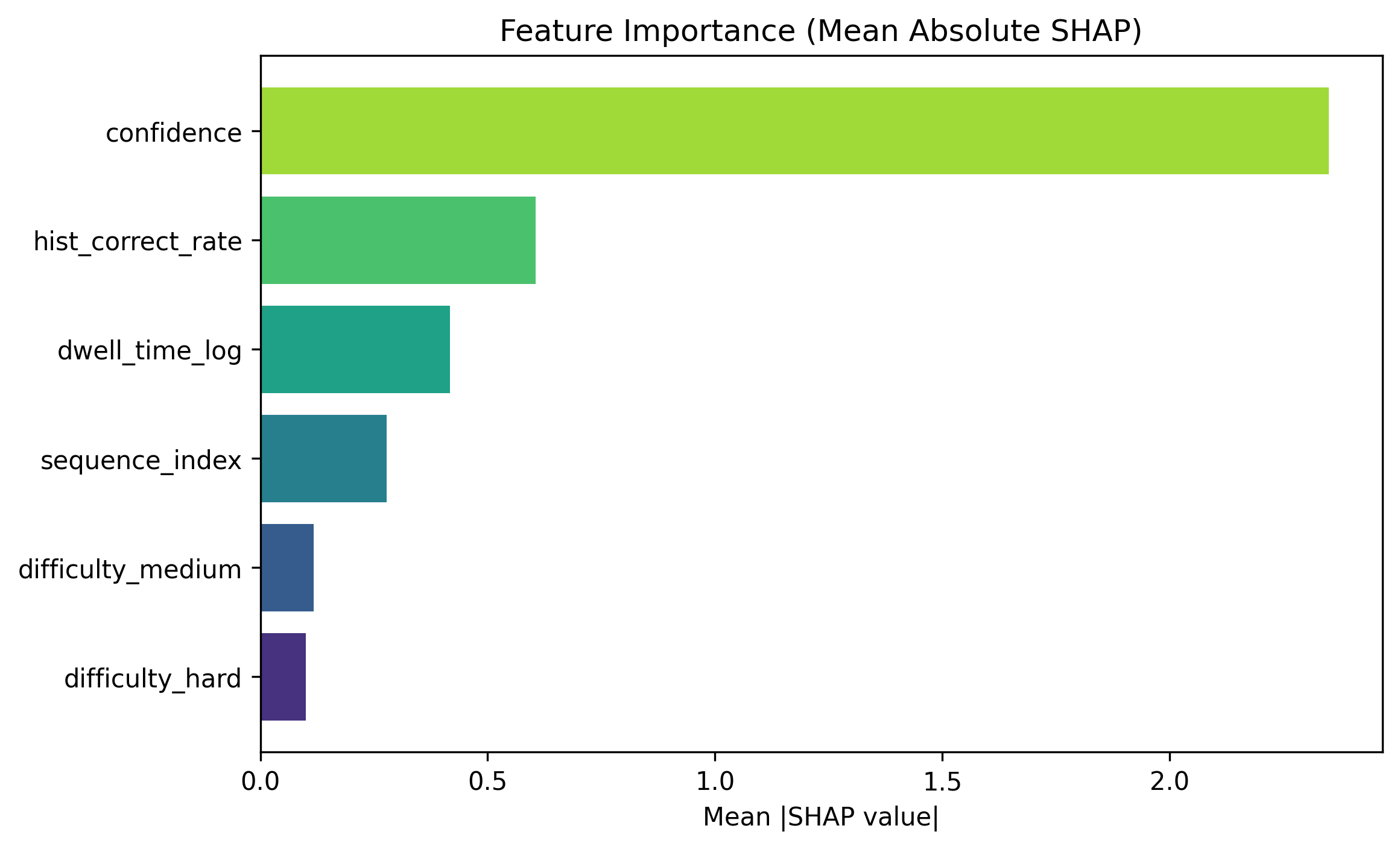}
\caption{Global SHAP feature importance (mean absolute SHAP values) for the LightGBM classifier. Confidence accounts for 60.8\% of total importance, with behavioral telemetry features (dwell time, sequence index) contributing 10.8\% and 7.2\% respectively. Difficulty features contribute minimally ($<$6\% combined).}
\label{fig:shap_bar}
\end{figure}

The SHAP beeswarm plot in Figure~\ref{fig:shap_beeswarm} and global importance bar chart in Figure~\ref{fig:shap_bar} demonstrate that student self-reported confidence is the dominant predictor of correctness, accounting for 60.8\% of the predictive weight. Historical correctness rate and log dwell time contribute 15.6\% and 10.8\% respectively, while task difficulty and sequence index play secondary roles.

\subsection{Ablation Study on Confidence Telemetry}
To evaluate the independent predictive signal in behavioral telemetry, I conduct an ablation study removing the self-reported confidence feature. The results are summarized in Table~\ref{tab:ablation_results}.

\begin{table}[!htbp]
\caption{Confidence Ablation Study Results}
\label{tab:ablation_results}
\centering
\begin{tabular}{lcccc}
\toprule
Condition & AUC & F1 & Brier & Acc \\
\midrule
Full Features & 0.878 & 0.789 & 0.147 & 0.789 \\
Ablated (No Conf.) & 0.649 & 0.612 & 0.224 & 0.628 \\
\midrule
\textbf{Delta ($\Delta$)} & \textbf{--0.229} & \textbf{--0.177} & \textbf{+0.077} & \textbf{--0.161} \\
\bottomrule
\end{tabular}
\end{table}

\begin{figure}[!htbp]
\centering
\includegraphics[width=0.75\columnwidth]{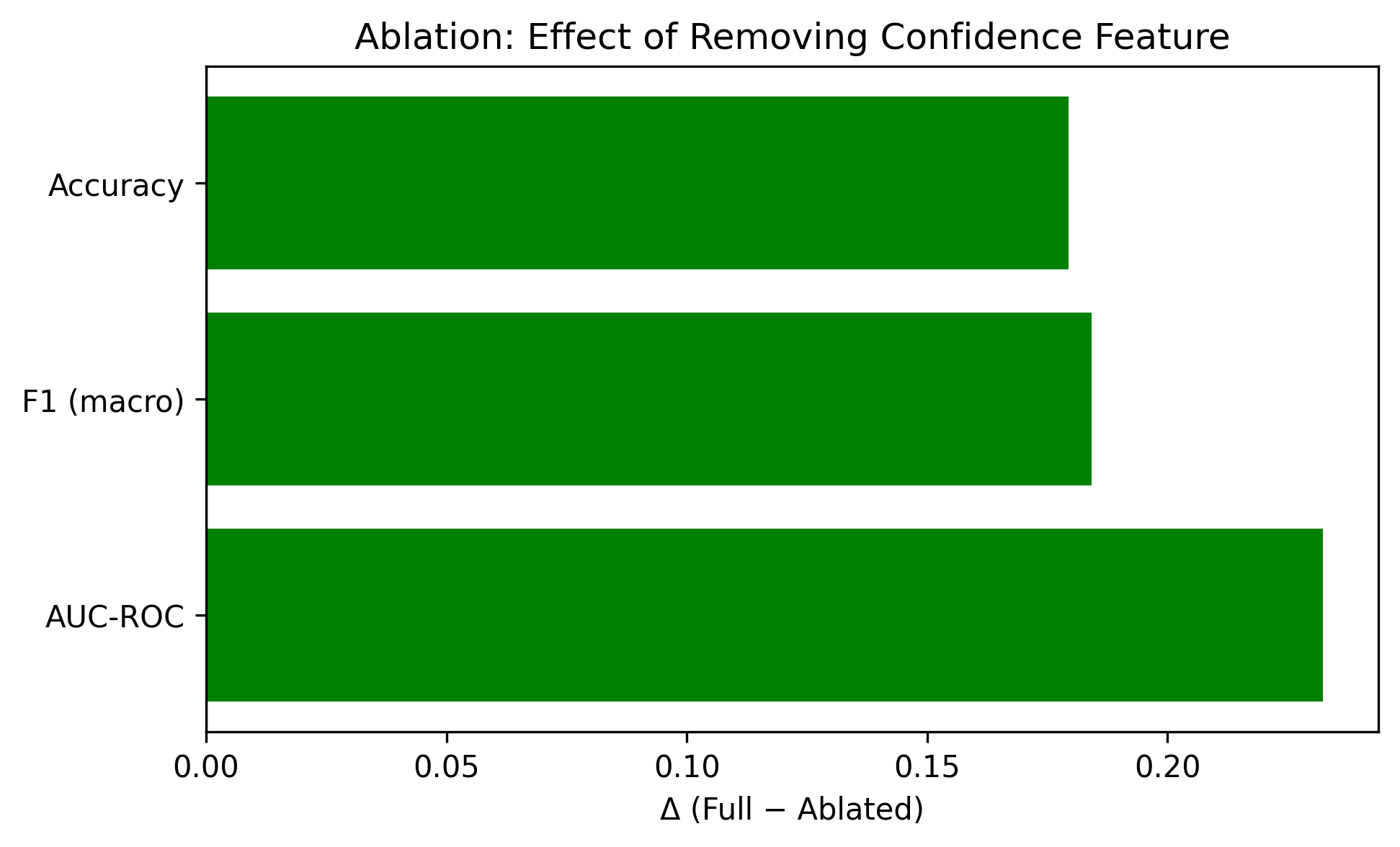}
\caption{Ablation study results: AUC-ROC comparison between the full-feature LightGBM model (AUC = 0.878) and the confidence-ablated model (AUC = 0.649), yielding $\Delta$AUC = 0.232. Removing self-reported confidence causes a catastrophic performance drop, confirming that confidence is the dominant predictive signal in this behavioral telemetry dataset.}
\label{fig:ablation_delta}
\end{figure}

Removing the self-reported confidence feature causes a catastrophic performance drop: AUC-ROC falls by 0.229 to 0.649, and classification accuracy decreases by 16.1\% (Table~\ref{tab:ablation_results}). This demonstrates that in this dataset, pre-execution behavioral telemetry has very limited independent predictive capability, and the predictive classifier is predominantly a proxy for student self-assessment. Figure~\ref{fig:ablation_delta} illustrates the drop in AUC-ROC.

\subsection{Metacognitive Calibration Comparison (H2)}
Module B evaluates the probability calibration of the student's self-reported confidence compared to the LightGBM classifier. The results are reported in Table~\ref{tab:calibration_comparison} and decomposed in Table~\ref{tab:brier_decomposition}.

\begin{table}[!htbp]
\caption{Calibration Metrics Comparison}
\label{tab:calibration_comparison}
\centering
\begin{tabular}{lcc}
\toprule
Metric & Student (Na\"{i}ve) & LightGBM \\
\midrule
ECE & 0.109 [0.087, 0.137] & 0.068 [0.049, 0.099] \\
MCE & 0.199 & 0.123 \\
Brier Score & \textbf{0.145} & 0.147 \\
\bottomrule
\end{tabular}
\end{table}

\begin{figure}[!htbp]
\centering
\includegraphics[width=0.85\columnwidth]{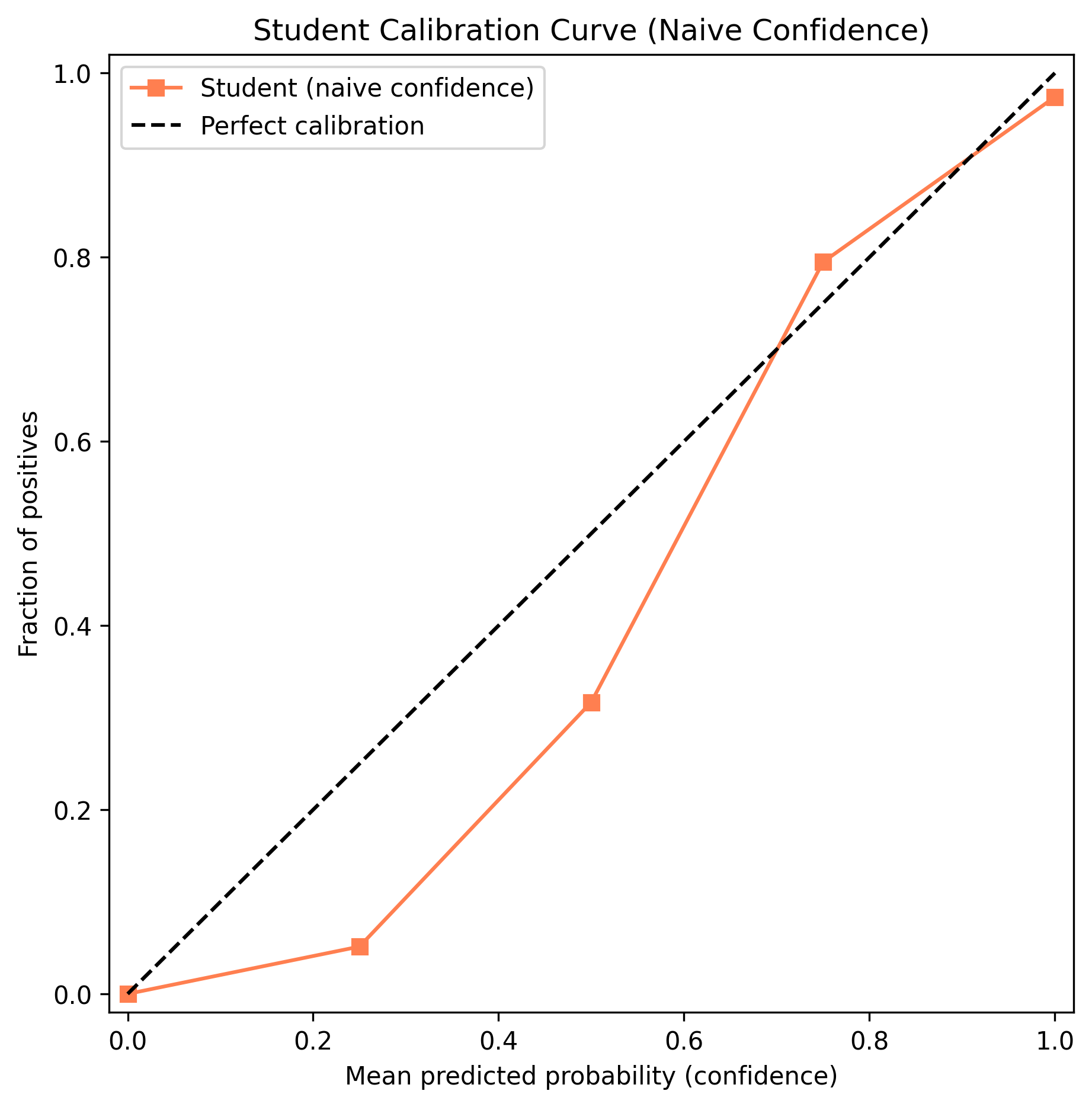}
\caption{Student metacognitive calibration curve mapping Likert confidence (scaled to $[0,1]$) against empirical accuracy. The dashed diagonal represents perfect calibration. The student Expected Calibration Error (ECE = 0.109) indicates systematic miscalibration, with overconfidence at higher confidence bins and underconfidence at lower bins.}
\label{fig:calibration_student}
\end{figure}

\begin{figure}[!htbp]
\centering
\includegraphics[width=0.85\columnwidth]{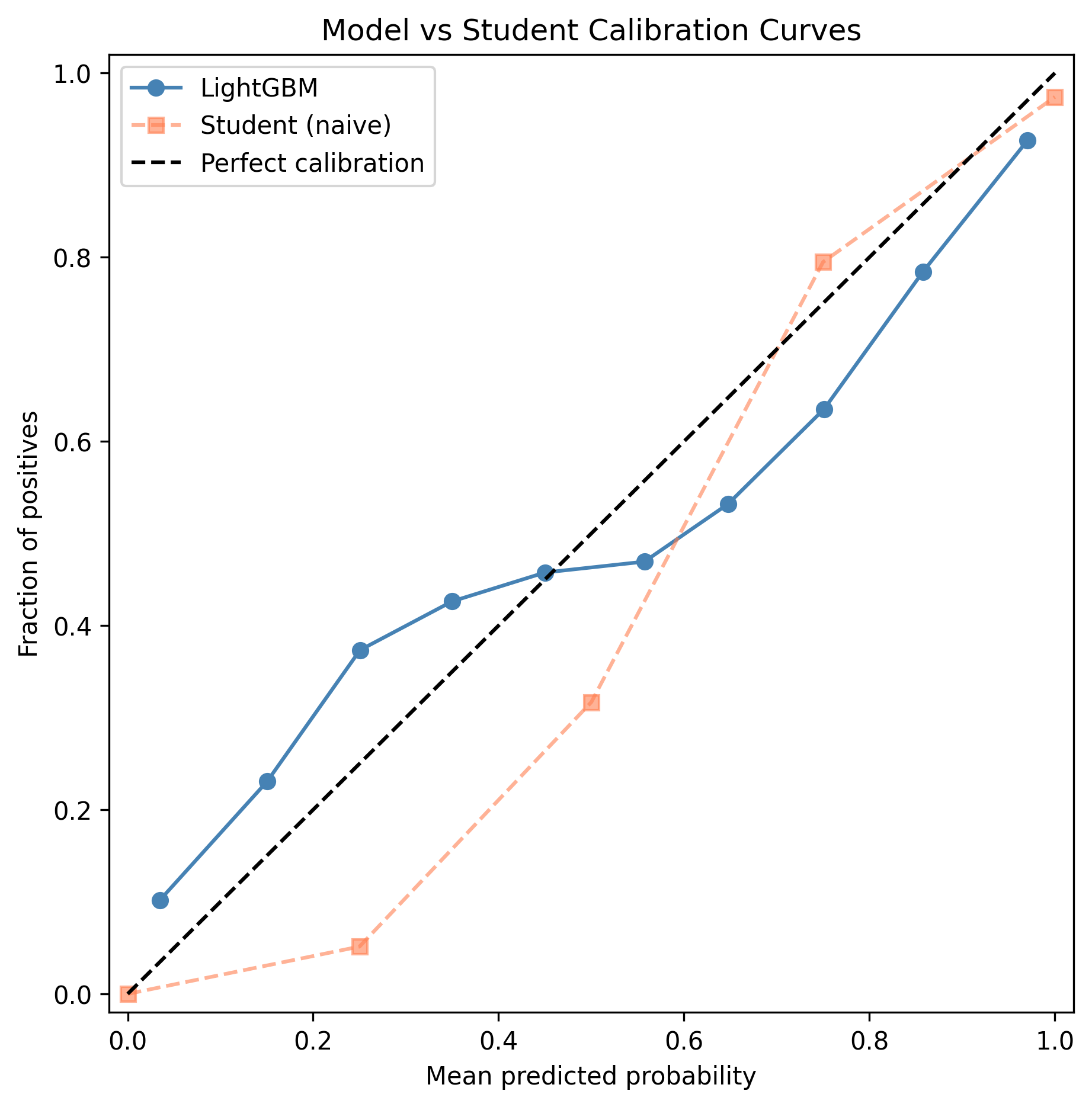}
\caption{LightGBM model calibration curve. The model's predicted probabilities are better aligned with empirical frequencies (ECE = 0.068) compared to student self-reported confidence (ECE = 0.109, Figure~\ref{fig:calibration_student}), confirming that the statistical model achieves superior probability calibration.}
\label{fig:calibration_model}
\end{figure}

\textbf{H2 is supported.} Student na\"{i}ve ECE (0.109) exceeds model ECE (0.068) by 0.041. The bootstrap confidence intervals show minimal overlap (student CI lower = 0.087, model CI upper = 0.099), demonstrating a statistically robust calibration gap. Students exhibit systematic overconfidence in high Likert bins and underconfidence in low bins (Figure~\ref{fig:calibration_student}). The model's calibration curve is shown in Figure~\ref{fig:calibration_model}.

\begin{figure}[!htbp]
\centering
\includegraphics[width=0.85\columnwidth]{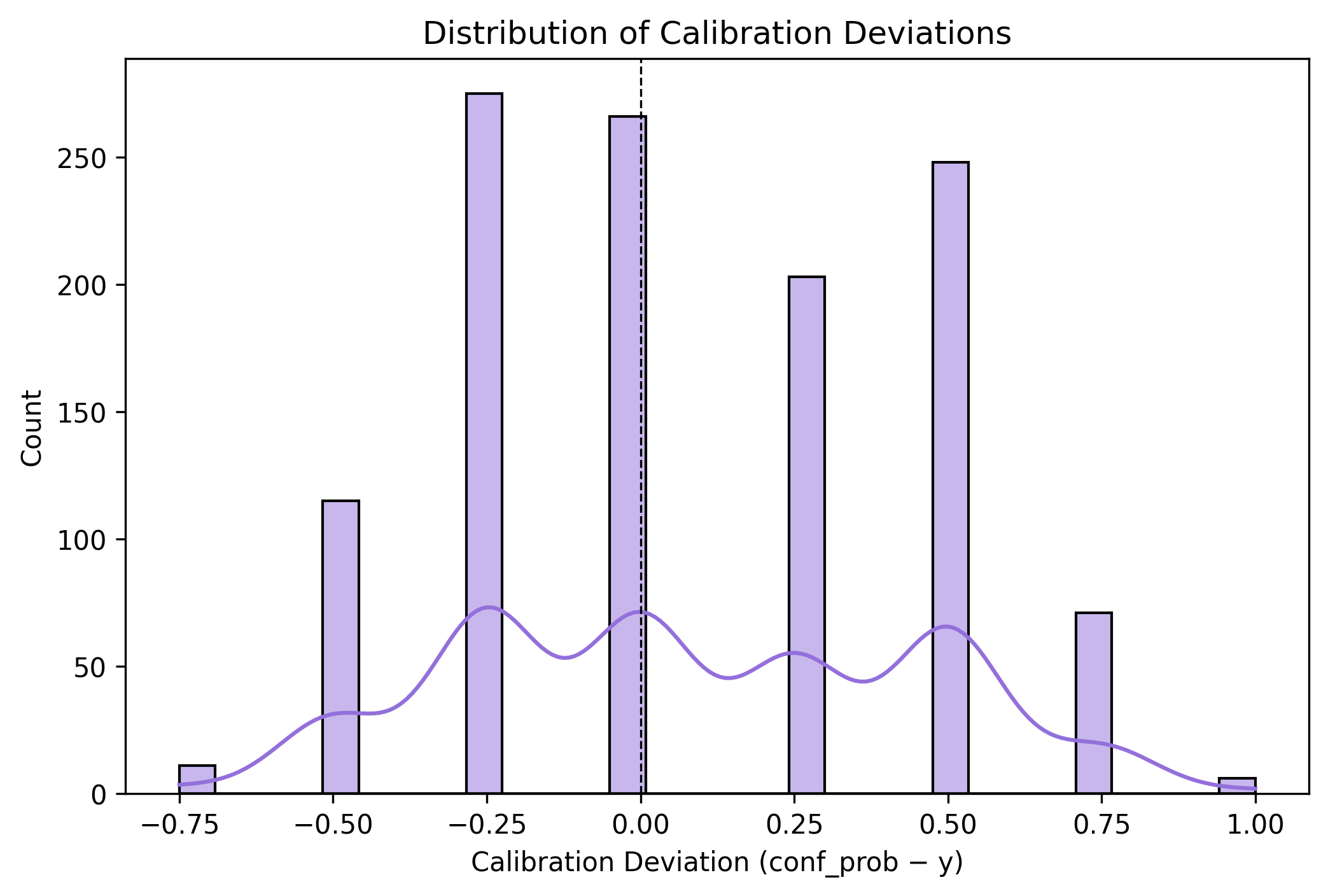}
\caption{Distribution of per-interaction calibration deviations $D_{ijk} = \text{conf}_{ijk} - y_{ijk}$. Positive values indicate overconfidence; negative values indicate underconfidence. The distribution is approximately symmetric around the mean, with high variance reflecting the dominance of interaction-specific (residual) factors over stable student- or item-level effects.}
\label{fig:calibration_deviation_dist}
\end{figure}

The distribution of per-interaction calibration deviations $D_{ijk}$ is illustrated in Figure~\ref{fig:calibration_deviation_dist}.

To inspect this further, I decompose the Brier score into Reliability, Resolution, and Uncertainty.

\begin{table}[!htbp]
\caption{Brier Score Decomposition}
\label{tab:brier_decomposition}
\centering
\begin{tabular}{lcc}
\toprule
Component & Student & LightGBM \\
\midrule
Reliability & 0.0180 & 0.0055 \\
Resolution & 0.1228 & 0.1069 \\
Uncertainty & 0.2494 & 0.2494 \\
\midrule
\textbf{Total Brier Score} & \textbf{0.145} & \textbf{0.147} \\
\bottomrule
\end{tabular}
\end{table}

The decomposition reveals a classic calibration-discrimination trade-off. The LightGBM model achieves dramatically lower reliability error (0.0055 vs. 0.0180), confirming that its predicted probabilities are better calibrated. However, the student's confidence achieves higher resolution (0.1228 vs. 0.1069), indicating that student self-assessment is slightly better at separating correct from incorrect attempts. This resolution advantage almost exactly offsets the model's calibration advantage, resulting in nearly identical overall Brier scores (0.145 vs. 0.147).

\subsection{Variance Decomposition of Calibration Deviation (H3)}
I fit the Crossed LMM to the calibration deviation $D_{ijk}$ using the full dataset. The variance components and Intra-class Correlation Coefficients (ICCs) are reported in Table~\ref{tab:variance_results}.

\begin{table}[!htbp]
\caption{Crossed GLMM Variance Components}
\label{tab:variance_results}
\centering
\begin{tabular}{lcc}
\toprule
Component & Variance ($\sigma^2$) & Proportion \\
\midrule
Student ($\sigma^2_u$) & 0.0179 & 12.3\% \\
Item ($\sigma^2_w$) & 0.0017 & 1.2\% \\
Residual ($\sigma^2_e$) & 0.1256 & 86.5\% \\
\midrule
\textbf{Total} & \textbf{0.1452} & \textbf{100.0\%} \\
\bottomrule
\end{tabular}
\end{table}

The resulting ICCs are:
\begin{itemize}
    \item $ICC_{\text{Student}} = 0.123$ (12.3\% of calibration variance is driven by individual student traits).
    \item $ICC_{\text{Item}} = 0.012$ (1.2\% of calibration variance is driven by item-specific characteristics).
\end{itemize}

\begin{figure}[!htbp]
\centering
\includegraphics[width=0.7\columnwidth]{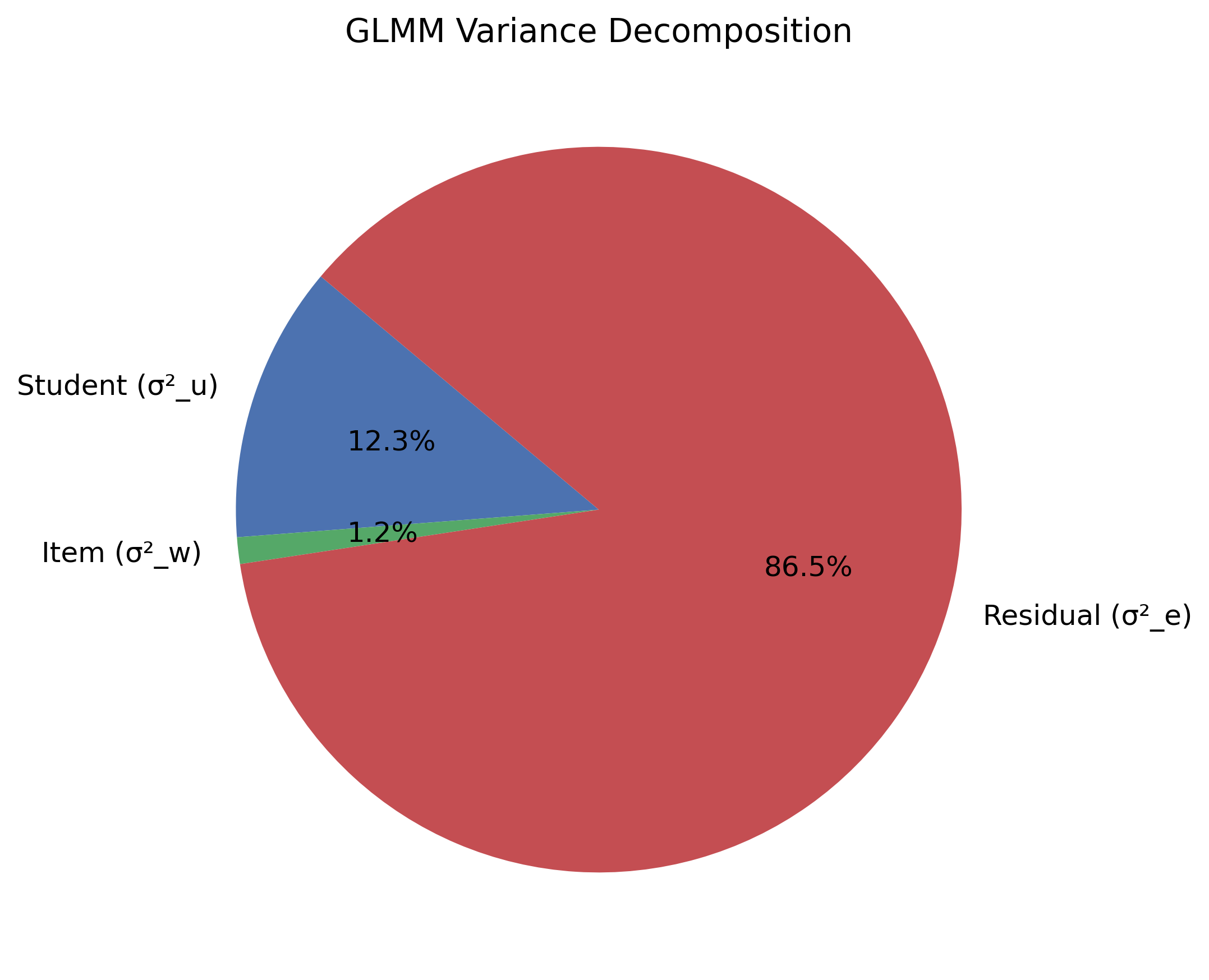}
\caption{Variance decomposition of calibration deviation from the crossed GLMM. Residual (interaction-specific) variance dominates at 86.5\%, with student-level variance contributing 12.3\% (ICC\textsubscript{Student} = 0.123) and item-level variance contributing only 1.2\% (ICC\textsubscript{Item} = 0.012). This indicates that metacognitive calibration is predominantly situational rather than dispositional.}
\label{fig:icc_variance}
\end{figure}

\textbf{H3 is not supported.} The student-level ICC (0.123) falls well below the pre-registered threshold of 0.20. Individual student traits account for only ${\sim}$12\% of the variation in calibration deviation, whereas 86.5\% of the variance is interaction-specific residual variance (Figure~\ref{fig:icc_variance}). This indicates that metacognitive calibration in this population is highly situational rather than dispositional.

The estimated fixed effects for the three behavioral predictors are reported in Table~\ref{tab:fixed_effects}.

\begin{table}[!htbp]
\caption{GLMM Fixed Effects Estimates}
\label{tab:fixed_effects}
\centering
\begin{tabular}{lcccc}
\toprule
Predictor & $\beta$ & SE & $z$ & $p$ \\
\midrule
Intercept & --0.133 & 0.083 & --1.61 & 0.107 \\
Log Dwell Time ($DT'$) & \textbf{0.046} & 0.016 & 2.94 & \textbf{0.003} \\
Sequence Index ($SI$) & 0.064 & 0.035 & 1.80 & 0.073 \\
Hist. Correctness ($HCR$) & 0.051 & 0.092 & 0.55 & 0.581 \\
\bottomrule
\end{tabular}
\end{table}

Log-transformed dwell time is the only statistically significant predictor of calibration deviation ($\beta_1 = 0.046$, $p = 0.003$), indicating that longer dwell times are associated with increased overconfidence. Sequence index exhibits a weak trend toward overconfidence over time ($\beta_2 = 0.064$, $p = 0.073$), while historical correctness rate has no significant fixed effect on calibration deviation ($p = 0.581$).

\subsection{Predictive-Explanatory Divergence Analysis (H4)}
I compute the PEDI between the normalized SHAP global feature importances $\mathbf{s}$ (Module A) and the standardized fixed-effects coefficients $\mathbf{v}$ (Module C) across the three shared features: Dwell Time ($DT'$), Sequence Index ($SI$), and Historical Correctness ($HCR$). The vectors are:
\begin{itemize}
    \item SHAP importance: $\mathbf{s} = [0.417, 0.278, 0.605]$
    \item Standardized GLMM weights: $\mathbf{v} = [0.046, 0.064, 0.051]$
    \item Normalized: $\mathbf{s}_{\text{norm}} = [0.321, 0.214, 0.465]$, $\mathbf{v}_{\text{norm}} = [0.286, 0.398, 0.317]$.
\end{itemize}

The observed PEDI values, bootstrap confidence intervals, and permutation $p$-values are reported in Table~\ref{tab:pedi_results}.

\begin{table}[!htbp]
\caption{Predictive-Explanatory Divergence Index (PEDI)}
\label{tab:pedi_results}
\centering
\begin{tabular}{lccc}
\toprule
Metric & Value & 95\% CI & $p$ \\
\midrule
$PEDI_{cos}$ & 0.081 & [0.060, 0.104] & 0.327 \\
$PEDI_{JS}$ & 0.022 & [0.016, 0.029] & 0.327 \\
\bottomrule
\end{tabular}
\end{table}

\begin{figure}[!htbp]
\centering
\includegraphics[width=0.85\columnwidth]{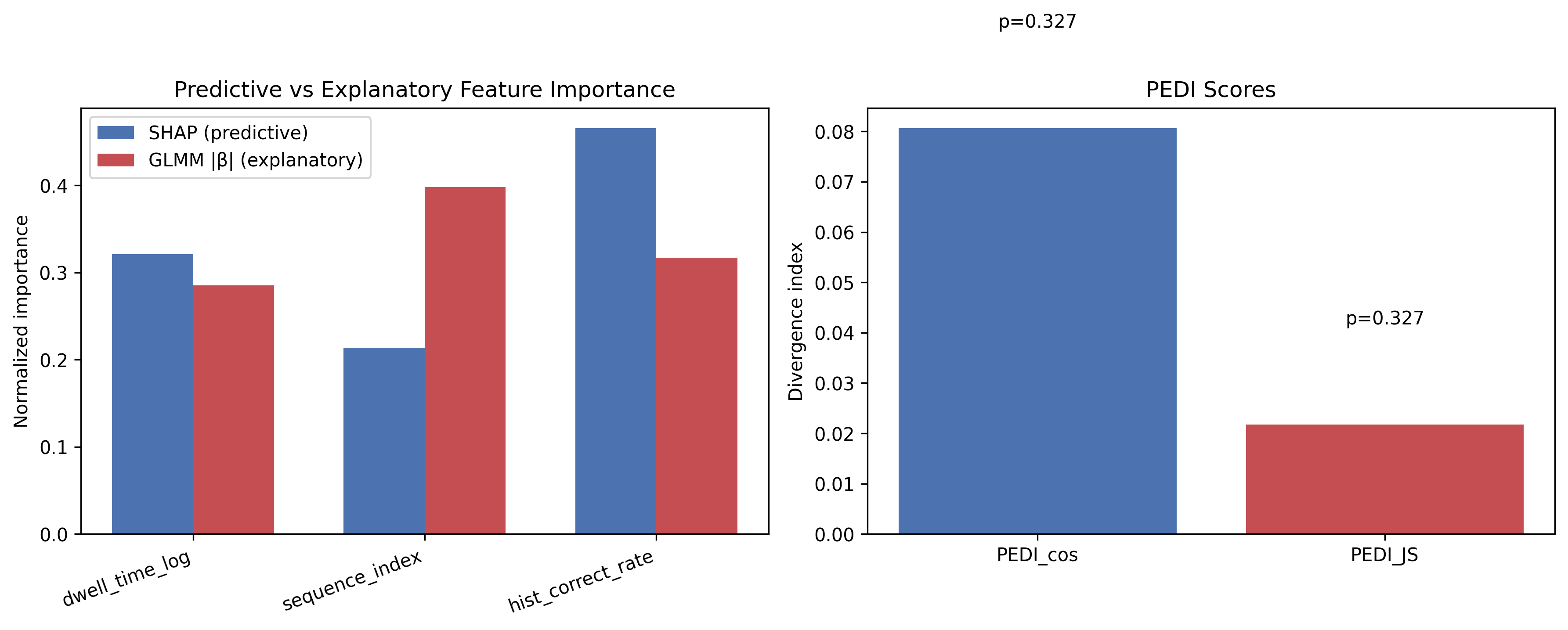}
\caption{Predictive-Explanatory Divergence Index (PEDI) values with 95\% bootstrap confidence intervals. PEDI\textsubscript{cos} = 0.081 [0.060, 0.104] and PEDI\textsubscript{JS} = 0.022 [0.016, 0.029]. Neither metric achieves statistical significance under permutation testing ($p$ = 0.327), indicating that SHAP feature importance and GLMM fixed-effect profiles are not significantly divergent across the three shared features.}
\label{fig:pedi_comparison}
\end{figure}

\begin{figure}[!htbp]
\centering
\includegraphics[width=0.85\columnwidth]{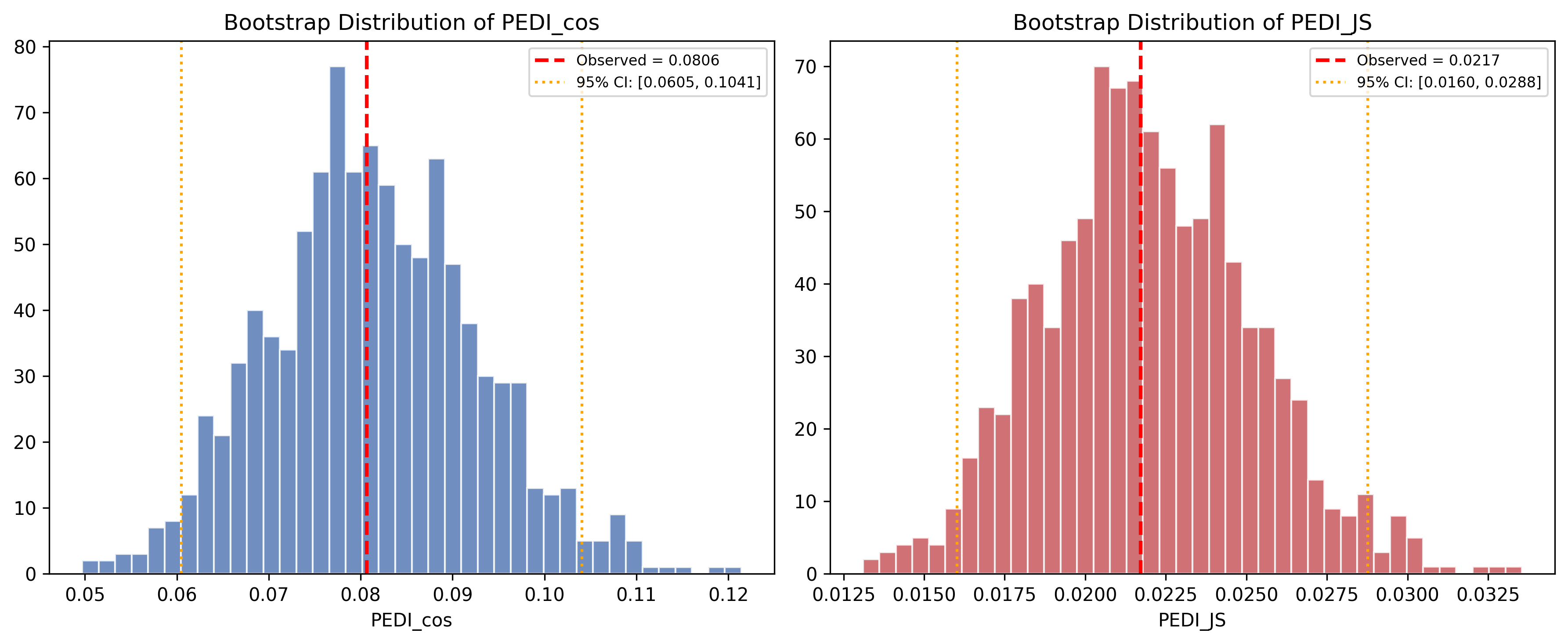}
\caption{Bootstrap distribution of PEDI\textsubscript{cos} (1,000 iterations) overlaid with the permutation null distribution. The observed divergence (vertical dashed line, PEDI\textsubscript{cos} = 0.081) falls within the null distribution, yielding $p$ = 0.327 and confirming that the predictive--explanatory divergence is not statistically significant.}
\label{fig:pedi_bootstrap}
\end{figure}

\textbf{H4 is not supported.} Neither variant of the PEDI achieves statistical significance ($p = 0.327$, Table~\ref{tab:pedi_results}). The observed cosine divergence ($0.081$) is small and falls well within the null permutation distribution (Figure~\ref{fig:pedi_comparison} and Figure~\ref{fig:pedi_bootstrap}). This indicates that for the shared behavioral features, the feature importance profile driving correctness prediction is structurally congruent with the fixed effects driving calibration deviation.

\section{Discussion}
The evaluation of UBP-CAP on student pre-execution telemetry yields several key methodological insights, particularly regarding the hypotheses that were not supported.

\subsection{The Confidence Dominance Problem and Feature Redundancy}
The primary finding from both SHAP feature attribution and the ablation study is the overwhelming predictive power of student self-reported confidence. Confidence accounts for over 60\% of the predictive weight in the LightGBM classifier and is responsible for a massive gap in AUC-ROC (0.878 with confidence vs. 0.649 without confidence). This dominance reveals a structural limitation in UBP-CAP: the classification model is predominantly predicting student correctness by utilizing the student's own self-reported confidence. Because confidence is the core variable used to define the calibration deviation ($D_{ijk} = conf_{ijk} - y_{ijk}$) in Module C, it was mathematically necessary to exclude confidence from the crossed random-effects model to avoid self-referential collinearity. 

Consequently, the PEDI consistency check compared only the three remaining behavioral features (dwell time, sequence index, and historical correctness). This confidence exclusion means that PEDI is only comparing the ``tail'' of the predictive importance distribution. This structural mismatch explains the lack of significance for H4: since the primary predictive signal (confidence) is omitted, the remaining features carry very low predictive signal, and their relative importances in both models are roughly congruent because they are both capturing residual behavioral noise.

\subsection{Calibration is Situational, Not Dispositional}
The crossed GLMM estimated $ICC_{\text{Student}} = 0.123$, which is substantially below the pre-registered threshold of 0.20. In educational research, student metacognition is often conceptualized as a stable individual trait---that is, some students are consistently overconfident, while others are consistently underconfident. My results challenge this trait-like assumption. Only 12\% of the variance in calibration deviation is driven by stable individual differences between students. Instead, the overwhelming majority (86.5\%) is residual variance, meaning that calibration deviations fluctuate dramatically from interaction to interaction for the same student. 

This situational nature has significant practical implications for intelligent tutoring systems. If calibration error is not a stable student trait, generic student-level interventions (e.g., warning a student that they tend to be overconfident) are unlikely to be effective. Instead, metacognitive scaffolding must be situational, providing item-level feedback that dynamically recalibrates confidence based on task-specific characteristics or real-time behavioral telemetry, such as dwell time.

\subsection{Simplicity Wins: Model Complexity on Small Datasets}
A notable finding is that Logistic Regression achieves the highest overall AUC-ROC (0.903), outperforming both Random Forest (0.891) and the gradient-boosted LightGBM model (0.878). Although these differences are not statistically significant, they highlight a common pitfall in educational data mining: defaulting to complex, non-linear ensemble models. On relatively small datasets (such as the 1,195 interactions), complex tree-based classifiers are prone to overfitting on student-level groupings, despite group-aware cross-validation. Simple linear models with L2 regularization provide robust, generalizeable decision boundaries, suggesting that linear relationships are sufficient to capture the main effects in pre-execution telemetry.

\subsection{The Significance of Dwell Time}
Log dwell time was the only significant behavioral fixed effect predictor of calibration deviation ($\beta = 0.046, p = 0.003$). The positive coefficient indicates that longer dwell times are associated with higher overconfidence. This is consistent with a ``deliberation-bias'' hypothesis: when a student spends an unusually long time viewing a task before submitting, their self-reported confidence increases, perhaps due to a subjective sense of effort (a ``sunk cost'' metacognitive bias). However, this extra deliberation time does not translate to a proportional increase in correctness, leading to an inflated calibration gap. This represents a valuable behavioral marker that tutoring systems can monitor in real time.

\section{Limitations}

\subsection{Sample Size and Power Constraints}
The primary limitation of this study is the sample size (27 students, 45 tasks, 1,195 interaction records). While sufficient for fitting mixed models, the small number of shared behavioral features (3 features) severely limited the statistical power of the PEDI permutation test. With only 3 dimensions, there are only $3! = 6$ possible feature label permutations, making it mathematically impossible to achieve a $p$-value lower than $1/6 \approx 0.167$ for a one-tailed test (or $2/6 = 0.333$ for a two-tailed test, matching the observed $p = 0.327$). Thus, the non-significance of H4 is partly a mathematical artifact of the feature set size.

\subsection{Statistical Approximations}
The Crossed GLMM was estimated using a two-stage MixedLM approximation in Python, as lme4-based packages failed to converge on the crossed random intercepts. While this approximation is standard, it may underestimate the variance components, slightly biasing the ICCs downward and inflating the residual variance. Future work should utilize Bayesian Markov Chain Monte Carlo (MCMC) estimation to obtain exact posterior distributions for the variance components.

\subsection{No Temporal Modeling}
The current UBP-CAP framework treats interactions as exchangeable within GroupKFold folds, ignoring the longitudinal sequence of tasks. Students likely learn and adapt their metacognitive calibration over the course of a session. Modeling these dynamic changes using state-space models or recurrent networks is a crucial direction for future work.

\section{Conclusion}
I have presented the Unified Behavioral Prediction and Calibration Analysis Pipeline (UBP-CAP), a novel methodological framework integrating machine learning correctness prediction, formal probability calibration evaluation, and crossed random-effects variance decomposition. By evaluating UBP-CAP on student pre-execution telemetry, I demonstrated that while correctness can be predicted with high accuracy (AUC-ROC = 0.903), the predictive model is heavily dominated by self-reported confidence. The student calibration analysis confirmed systematic metacognitive miscalibration (ECE = 0.109 vs. model ECE = 0.068). Importantly, variance decomposition showed that metacognitive calibration is situational (ICC\textsubscript{Student} = 0.123), suggesting that intelligent tutoring systems should focus on task-specific, real-time feedback rather than student-level traits. Finally, the Predictive-Explanatory Divergence Index (PEDI) provided a rigorous protocol to contrast predictive and explanatory models, revealing no significant structural divergence on the shared behavioral features. Future work will extend UBP-CAP to larger longitudinal datasets and incorporate temporal sequence models.

\bibliographystyle{IEEEtran}
\bibliography{I8}

\end{document}